\documentclass[lettersize,journal]{IEEEtran}
\usepackage{amsmath,amsfonts}
\usepackage{algorithmic}
\usepackage{algorithm}
\usepackage{array}
\usepackage[caption=false,font=normalsize,labelfont=sf,textfont=sf]{subfig}
\usepackage{textcomp}
\usepackage{stfloats}
\usepackage{url}
\usepackage{verbatim}
\usepackage{graphicx}
\usepackage{cite}
\usepackage[hidelinks]{hyperref}
\usepackage{color}
\usepackage{siunitx}
\usepackage{multirow}
\usepackage{microtype}
\usepackage{booktabs} 
\usepackage{caption}
\begin{document}

\title{MLM: Learning Multi-task Loco-Manipulation Whole-Body Control for Quadruped Robot with Arm}

\author{Xin Liu$^{1}$, Bida Ma$^{1}$, Chenkun Qi$^*$$^{1}$,~\IEEEmembership{Senior Member,~IEEE}, Yan Ding$^{\dagger}$$^{2}$, Nuo Xu$^{1}$, Zhaxizhuoma$^{2}$, Guorong Zhang$^{2}$, Pengan Chen$^{2}$, Kehui Liu$^{2}$, Zhongjie Jia$^{2}$, Chuyue Guan$^{2}$, Yule Mo$^{1}$, Jiaqi Liu$^{1}$, Feng Gao$^{1}$, Jiangwei Zhong$^{3}$, Bin Zhao$^{2}$, and Xuelong Li$^{4}$,~\IEEEmembership{Fellow,~IEEE}
\thanks{This work was supported by the National Natural Science Foundation of China (Grant No. 52575029). (Corresponding author:
Chenkun Qi and Yan Ding.)}
\thanks{$^{1}$ Xin Liu, Bida Ma, Chenkun Qi, Nuo Xu, Yule Mo, Jiaqi Liu, and Feng Gao are with School of Mechanical Engineering, Shanghai Jiao Tong University, Shanghai, 200240, China (e-mail: liu15764167516@sjtu.edu.cn, mabida@sjtu.edu.cn; chenkqi@sjtu.edu.cn; noraxu@sjtu.edu.cn; moyule.com@sjtu.edu.cn; liujiaqi443@sjtu.edu.cn; fengg@sjtu.edu.cn).}
\thanks{$^{2}$ Yan Ding, Zhaxizhuoma, Guorong Zhang, Pengan Chen, Kehui Liu, Zhongjie Jia, Chuyue Guan, and Bin Zhao are with Shanghai AI Laboratory, Shanghai, 200030, China (e-mail: yding25@binghamton.edu; zxzmkufufu@gmail.com; 15821192606@163.com; cpa2001@connect.hku.hk; kehuiliu@mail.nwpu.edu.cn; jiazhongjie@sjtu.edu.cn; ilovegcy@126.com
; binzhao111@gmail.com).}
\thanks{$^{3}$ Jiangwei Zhong is with Lenovo Corporation, Shanghai, 201203, China (e-mail: zhongjw@lenovo.com).}
\thanks{$^{4}$ Xuelong Li is with Institute of Artificial Intelligence (TeleAI), China Telecom, Shanghai, 200233, China (e-mail: li@nwpu.edu.cn).}
}



\maketitle

\begin{abstract}
Whole-body loco-manipulation for quadruped robots with arms remains a challenging problem, particularly in achieving multi-task control. To address this, we propose MLM, a reinforcement learning framework driven by both real-world and simulation data. It enables a six-DoF robotic arm--equipped quadruped robot to perform whole-body loco-manipulation for multiple tasks autonomously or under human teleoperation.
To address the problem of balancing multiple tasks during the learning of loco-manipulation, we introduce a trajectory library with an adaptive, curriculum-based sampling mechanism. This approach allows the policy to efficiently leverage real-world collected trajectories for learning multi-task loco-manipulation.
To address deployment scenarios with only historical observations and to enhance the performance of policy execution across tasks with different spatial ranges, we propose a Trajectory-Velocity Prediction policy network. It predicts unobservable future trajectories and velocities.
By leveraging extensive simulation data and curriculum-based rewards, our controller achieves whole-body behaviors in simulation and zero-shot transfer to real-world deployment. Ablation studies in simulation verify the necessity and effectiveness of our approach, while real-world experiments on a Go2 robot with an Airbot robotic arm demonstrate the policy's good performance in multi-task execution.
\end{abstract}

\section{Introduction}
\IEEEPARstart{Q}{uadruped} robot motion control, especially through reinforcement learning (RL), has gained significant attention and development, enhancing their diverse and impressive capabilities~\cite{lee2020learning, wjzamp, kumar2021rma, ji2022concurrent, Li2024Learning, Peng2025Learning}. 
A six-degree-of-freedom (DoF) robotic arm can offer more possibilities for enabling quadruped robots to perform manipulations. By integrating the locomotion of a quadruped robot with the manipulation capabilities of a robotic arm, a broader range of tasks can be accomplished~\cite{Sleiman2023Versatile, Yokoyama2024ASC}.
\begin{figure}[tbp]
\includegraphics[width=0.9\linewidth]{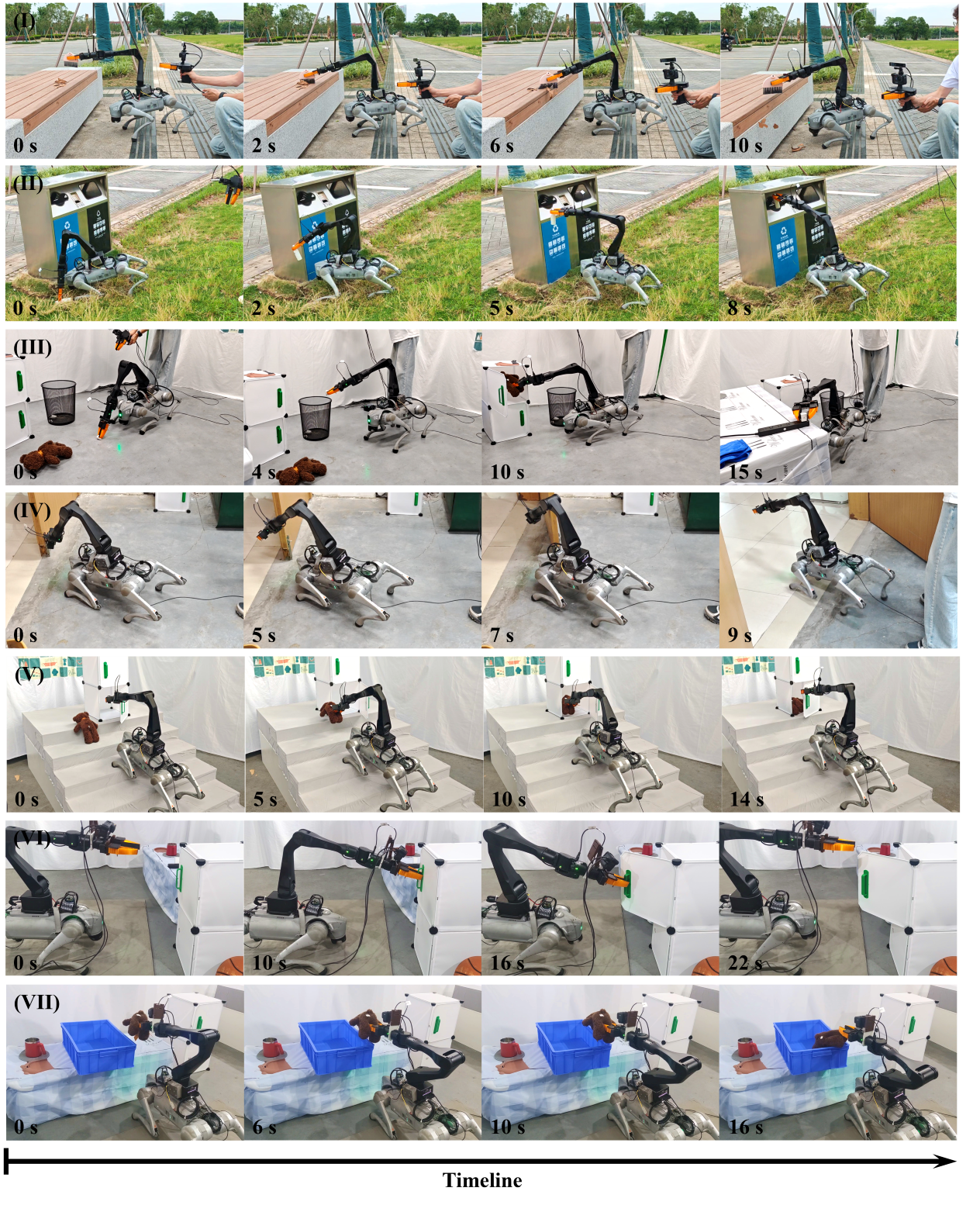}
\centering
\caption{The robot showcases its ability to perform multi-task whole-body loco-manipulation in the real world.}
\label{fig:real_robot}
\vspace{-15pt}
\end{figure}

Employing a single policy can reduce control complexity by eliminating the need for task-specific policies. However, achieving multiple loco-manipulation tasks with a single policy presents a challenge for quadruped robots, and balancing multiple tasks remains unresolved in existing methods, making an adaptive mechanism essential and warranting further research.
Moreover, in simulation-driven paradigms, task-specific assets and rewards demand heavy manual design with poor generalizability.
The visual gap for manipulation also remains challenging due to differences in texture, color, lighting, and other factors between simulation and reality~\cite{Mohammad2024SplatSim}.
A feasible solution is to distill multiple tasks into multiple trajectories learning. This avoids incorporating visual data of the task objectives into the control.

Building on this idea, real-world multi-task trajectories can be introduced into simulation, while simulation can generate large-scale interaction data. By abstracting tasks into trajectories and defining a unified reward, quadruped robots can be guided to develop coordinated whole-body behaviors under a single RL policy. Real-world multi-task data provide spatial trajectories consistent with physical dynamics and distributed across the robot’s work space, which are essential for learning whole-body behaviors and generalizing to unseen trajectories~\cite{He2025ASAP}. In contrast, random spatial curves sampled in simulation lack realistic dynamics such as velocity and acceleration.

The acquisition of trajectories can be achieved through various methods, including UMI-like approaches~\cite{chi2024umi, Zhaxizhuoma2025FastUMI}, teleoperation~\cite{Tony2023Learning, Anthony2023RT1}, exoskeleton-based systems~\cite{Fang2024AirExo}. 
It is important to note that we do not collect the whole-body trajectory data, which distinguishes our approach from imitation learning. This is due to the challenges of species selection for data collection and the inability to obtain such data from real quadruped robot with an arm. Thus, we use the UMI-like approach  to obtain only the end-effector 6D manipulation trajectories, a lightweight solution independent of the robots.

In deployment schemes that only have access to historical trajectories, such as teleoperation, future trajectory information cannot be used as observation. However, future information is crucial for the stability of whole-body execution tasks, as it indicates the direction of the next step in the action~\cite{Huy2024UMIonLegs}. Thus, a trajectory prediction mechanism is necessary.

To address the aforementioned challenges, we incorporate Fast-UMI~\cite{Zhaxizhuoma2025FastUMI} to provide real-world manipulation trajectories and introduce MLM, a learning-based multi-task loco-manipulation framework. The control policy is trained in simulation and subsequently transferred to the real world.
Our contributions are summarized as follows:
\begin{itemize}
\item We propose a Trajectory-Velocity Prediction-based RL framework for training a multi-task loco-manipulation controller, driven by both real-world and simulation data, with trajectory interface deployment enabled through teleoperation and a diffusion policy.
\item We introduce a multi-task real-world trajectory data library, where adaptive and curriculum-based sampling enables the robot to balance performance across tasks and learn multi-task manipulation with a single policy network.
\item We deploy a real quadruped robot with an arm and validate the excellent performance of our method in whole-body loco-manipulation control across multiple tasks.
\end{itemize}

\section{Related Work}
\subsection{RL-based Legged Locomotion}
The success of RL-based motion controllers for legged robots, compared to model-based methods, lies in two main advantages. First, RL trains control policies using large-scale simulated robot–environment interactions, capturing robot dynamics while leveraging far more data than real-world training allows~\cite{lee2020learning,kumar2021rma,miki2022learning}. Second, reward-driven policies enable end-to-end mapping from states to joint action, eliminating the need for costly optimization during inference~\cite{margolis2022rapid,ji2022concurrent}.

Built on the parallel simulator Isaac Gym, quadruped robots can learn to walk within minutes \cite{rudin2022learning}. To enable sim-to-real transfer, an network was used to capture the complex dynamics of the ANYmal robot's motors \cite{hwangbo2019learning}. This approach was later extended with privileged learning, yielding a blind locomotion controller for challenging terrains \cite{lee2020learning}. Further advances include RMA, which enhances robustness in quadruped locomotion \cite{kumar2021rma}. To accelerate gait learning and adapt to complex terrains, researchers introduced gait priors \cite{escontrela2022adversarial,wjzamp,Liu2024Skill}, leveraging flat-terrain motion data for faster convergence. These efforts have greatly advanced legged locomotion \cite{Fan2024Rethinking, David2024ANYmalparkour, Miki2024Learning3D, Tairan2024Agile}, yet the development of loco-manipulation remains largely unexplored.

\subsection{Legged Loco-Manipulation}
To enhance legged robots with manipulation skills, researchers have combined their mobility with the manipulation capabilities of robotic arms. 
For instance, in \cite{Bellicoso2019ALMA}, footstep planning was addressed, and a hierarchical optimization-based controller was developed to enable a quadruped robot with a six-DoF robotic arm to perform manipulation. The study in \cite{Sleiman2023Versatile} proposed a approach that searches for whole-body trajectories and determines contact schedules to solve mobile manipulation tasks within predefined environments. Nonetheless, these optimization-based approaches often require task-specific design, presenting significant challenges in integrating the execution of multiple tasks within a single control policy. Additionally, they incur substantial computational overhead due to the optimization processes involved.

In recent years, several studies have focused on employing RL to achieve end-to-end control of legged robots with robotic arms. 
In \cite{Ma2022Combining}, the arm’s motion is modeled as predictable external torques, which the RL-trained locomotion policy compensates for. However, this non-whole-body approach limits the arm from leveraging the quadruped’s mobility and posture adjustments to extend its manipulation range. 
To enable whole-body control, an RL-based regularized online adaptation method was proposed, successfully achieving sim-to-real transfer for quadruped loco-manipulation \cite{Fu2023DeepWholeBody}. Despite this, the arm-base-defined manipulation space is vulnerable to body shaking and thus unstable. To address this, we adopt a world frame, where operational objectives remain independent of body motion.
In \cite{Jiang2025Learning}, a nonlinear reward fusion module was proposed to achieve whole-body behaviors for a wheeled quadrupedal manipulator. However, it did not integrate automatic execution of the task.
In \cite{Liu2024VisualWholeBody}, an RL-based approach that integrates visual information for autonomous loco-manipulation execution was proposed. Yet, the inverse kinematics (IK) strategy is also susceptible to locomotion interference due to the arm base being affected by the body’s movement. 
In~\cite{Huy2024UMIonLegs}, a vision-driven, trajectory-based loco-manipulation in the task space was realized for a quadruped robot. However, it did not consider the integration of multiple tasks and the balance of performance across tasks within a single policy. Additionally, it is only applicable to trajectory generation methods that involve future trajectory reasoning, such as diffusion policy (DP). Teleoperation that only involve historical trajectories are not suitable.

\section{Whole-body Control Method }
\begin{figure*}[htbp]
\vspace{-1.0em}
\setlength{\abovecaptionskip}{0.cm}
\setlength{\belowcaptionskip}{-0.cm}
\centering
\includegraphics[width=0.77\linewidth]{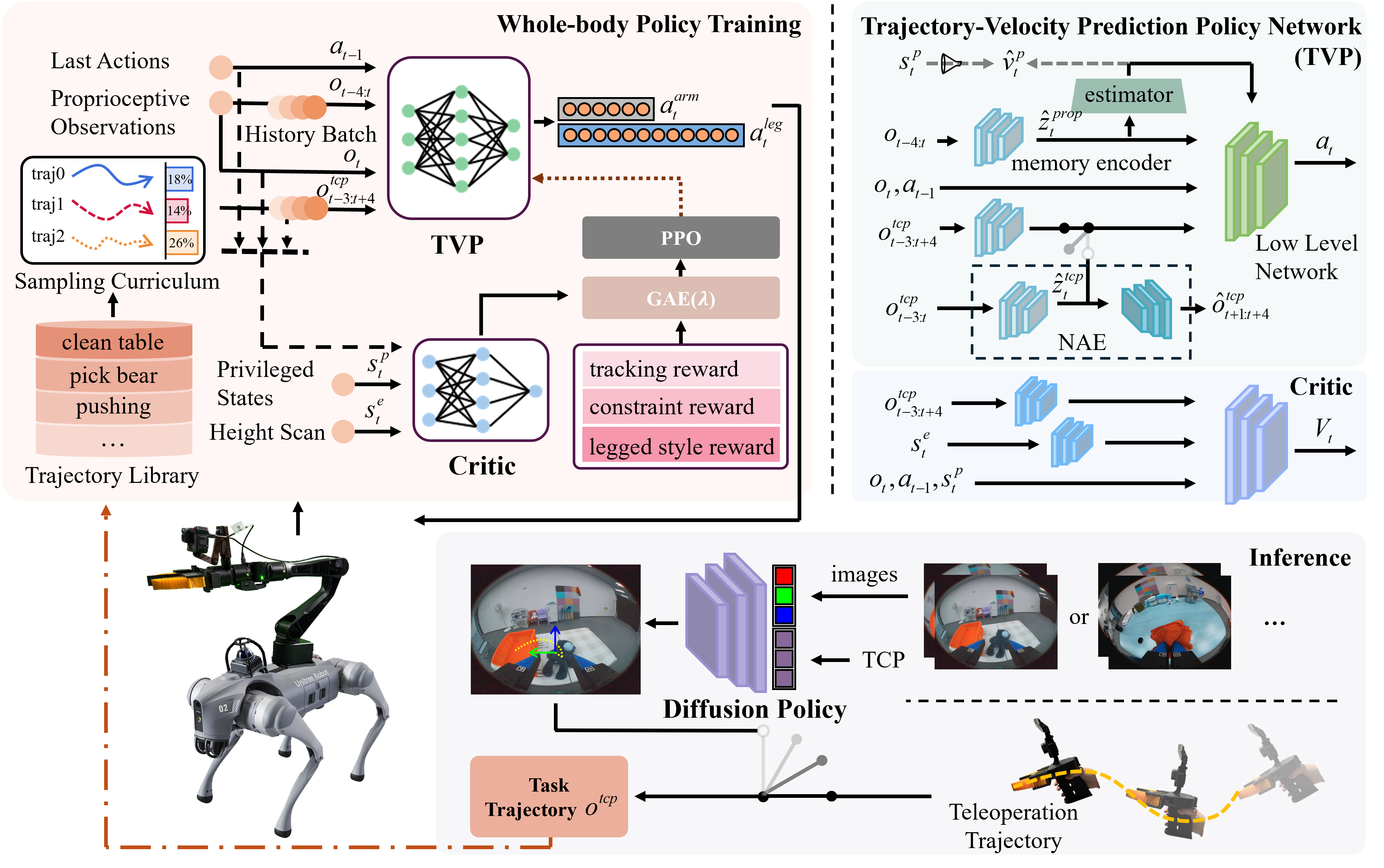}
\caption{Overview of the pipeline. We use RL to train a whole-body control policy. A trajectory library, together with an adaptive sampling curriculum, supplies multiple real-world manipulation trajectories for the policy to track. 
An NAE encodes historical trajectories to estimate future targets, while the policy can also take future trajectories directly, supporting two deployment modes: teleoperation and automatic DP. In addition, we train a supervised estimator to predict the base linear velocity $\hat{v} _{t}^{p}$.}
\label{fig:method}
\vspace{-14pt}
\end{figure*}

We develop a real-world and simulation dual data driven RL framework for quadruped robot whole-body loco-manipulation control. During training, a library containing real-world manipulation trajectories for multiple tasks is introduced, and an end-to-end Trajectory-Velocity Prediction policy network is trained for control. The pipeline is illustrated in Figure \ref{fig:method}.

\subsection{Whole-body Loco-Manipulation Policy}
We eliminate the requirement for object and environment perception in simulation, which accelerates training and avoids the large sim-to-real visual gap. Since information is partially observable, the loco-manipulation task is modeled as a partially observable Markov decision process (POMDP)~\cite{miki2022learning}. RL is employed to train a whole-body policy $\pi_{\theta}$, aiming to find optimal parameters $\theta$ that maximize the discounted expected return $\mathbb{E} _{\mathbf{\tau } ( \pi_{\theta} )  }  [ \sum_{t=0}^{T} {\gamma ^{t} r_{t} } ]$, where $r_t$ is the reward at time $t$, and $\gamma$ is the discount factor ($\gamma = 0.99$).
We introduce an asymmetric actor-critic framework to train the policy \cite{ji2022concurrent}. This approach is designed to address the POMDP problem while also avoiding the complexity of the two-stage training.

\textbf{State space:} 
The input of the actor (policy network) includes a five-frame history sequence of the quadruped robot’s proprioception $\boldsymbol{o}_t^q\in {\mathbb{R}^{30}}$, the six-DoF robotic arm’s proprioception $\boldsymbol{o}_t^a\in {\mathbb{R}^{12}}$, the last action $\boldsymbol{a}_{t-1}\in {\mathbb{R}^{18}}$, and the end-effector pose trajectory $o_{t-3:t+4}^{tcp} \in {\mathbb{R}^{72}}$. Besides the actor's input, the critic has access to the privileged state $\boldsymbol s_t^p\in{\mathbb{R}^{21}}$ and the terrain elevation data $\boldsymbol{s}_t^e\in{\mathbb{R}^{187}}$, which are only available during training and not used by the actor. A normalized autoencoder (NAE) only receives the historical end-effector pose sequence to predict future trajectory, while the memory encoder only receives the historical robot and arm proprioception. Further details are provided in Section~\ref{NETWORK}.

Specifically, $\boldsymbol{o}_t^q$ includes the quadruped's body angular velocity $\boldsymbol{\omega}_t^q\in {\mathbb{R}^{3}}$, the gravity vector $\boldsymbol{g}_t^q\in {\mathbb{R}^{3}}$ representing its orientation, leg joint positions $\boldsymbol{\theta}_t^q\in {\mathbb{R}^{12}}$ and velocities $\Dot{\boldsymbol{\theta}}_t^q\in {\mathbb{R}^{12}}$. The arm proprioception $\boldsymbol{o}_t^a$ contains the arm joint positions $\boldsymbol{\theta}_t^a\in {\mathbb{R}^{6}}$ and velocities $\Dot{\boldsymbol{\theta}}_t^a\in {\mathbb{R}^{6}}$. The last action $\boldsymbol{a}_{t-1}$ includes the previous time step’s leg joint action $\boldsymbol{a}_{t-1}^q\in {\mathbb{R}^{12}}$ and arm joint action $\boldsymbol{a}_{t-1}^a\in {\mathbb{R}^{6}}$. The trajectory pose sequence $o_{t-3:t+4}^{tcp}$ comprises Tool Center Points (TCP) over past and subsequent steps relative to time $t$, each including a 3D position and the first six elements of the rotation matrix. The privileged state $\boldsymbol s_t^p$, covering both robot and environment, includes the quadruped's body linear velocity $\boldsymbol{v}_t^q\in{\mathbb{R}^{3}}$, contact force $\boldsymbol{f}^{c}_t\in{\mathbb{R}^{12}}$, external force $\boldsymbol{f}^{e}_t\in{\mathbb{R}^{3}}$, and its position $\boldsymbol{p}^{e}_t\in{\mathbb{R}^{3}}$ on the robot. Finally, $\boldsymbol{s}_t^e$ represents the heights of 187 terrain points evenly distributed around the robot frame.

\textbf{Action space:} 
The whole-body policy outputs 12 joint position offsets for the legs, $\boldsymbol{a}_{t}^{q}\in{\mathbb{R}^{12}}$, and 6 joint position offsets for the arm, $\boldsymbol{a}_{t}^{a}\in{\mathbb{R}^{6}}$. These offsets are added to the default positions and sent to the motors as target position commands for proportional-derivative (PD) control \cite{Jiang2025Learning}.

\textbf{Reward:} 
Maintaining coordinated whole-body behaviors is challenging. Building on~\cite{Liu2024Skill}, we use adversarial motion prior (AMP) to encourage quadruped gaits learning via a style reward. During policy training, an adversarial discriminator is trained concurrently to provide the style reward, defined as $r^s = \max [0, 1-0.25\times\left(d^{\rm score}- 1\right)^2 ]$.

The primary reward of the policy, $r^{g}$, is the tracking of the robotic arm's TCP position and orientation:
\begin{equation} \label{tracking reward}
\begin{aligned}
r^{g} &= r^{g_p} \cdot r^{g_o} \\ 
r^{g_p} &= \exp({- (e_{1}^{p}/\sigma_{1}^{p} \oplus e_{2}^{p}/\sigma_{2}^{p} \oplus ... \oplus e_{N}^{p}/\sigma_{N}^{p})}) \\
r^{g_o} &= \exp ({- (e_{1}^{o}/\sigma_{1}^{o} \oplus e_{2}^{o}/\sigma_{2}^{o} \oplus ... \oplus e_{N}^{o}/\sigma_{N}^{o}) })
\end{aligned}
\end{equation}
where $N$ is the total number of tasks. `$\oplus$' represents the concatenation operation. We have established a curriculum to update different tasks reward factors $\sigma_{n}^{p}$ and $\sigma_{n}^{o}$, more details are described in~\ref{Trajectory_Sampling_Update}. The constraint reward, $r^{l}$, is used to regulate the robot's reasonable action, energy limitations, and hardware constraints. All rewards are shown in Table \ref{tab:reward details}.

\begin{table}[htbp]
        \renewcommand{\arraystretch}{1.0}
        \caption{Rewards for learning whole-body loco-manipulation.}
        \centering
        \label{tab:reward details}
        \begin{tabular}{@{}lll@{}}
        \toprule \\[-4mm]
        \textbf{Term}         & \textbf{Equation}       &\textbf{Weight}\\ \\[-4mm]\midrule
$r^{g}$  & (\ref{tracking reward}) & 2.0 \\ \midrule
\multirow{6}{*}{$r^{l}$}& $-\Vert\boldsymbol{\tau}\Vert_2$     & $1E-4$\\
                      & $-\Vert\boldsymbol{\ddot q}\Vert_2$      & $2.5E-7$\\
                      & $-\Vert \boldsymbol{a}_{t-1}-\boldsymbol{a}_t\Vert_2$ & 0.1\\
                      & $-n_{collision}$    & 0.1\\
                      &$-\Vert\max\left(\left|\boldsymbol{\tau} \right|-\boldsymbol{\tau}^{limit},0\right)\Vert_2$   & 0.05   \\
                              & $-\Vert\max\left(\left|\boldsymbol{\dot q} \right|-\boldsymbol{\dot q}^{limit},0\right)\Vert_2$     & 0.5  \\
                      \midrule
$r^s$                 & $\max [0, 1-0.25\times(d^{\rm score}- 1)^2 ]$  & 0.5\\
            \bottomrule
            \end{tabular}
\vspace{-0.5em}
\end{table}

\textbf{Terrain curriculum:} 
In addition to offering greater freedom of posture adjustment, the leg design of quadruped robots enables adaptation to terrains. To support loco-manipulation in such environments, we introduce a terrain curriculum \cite{Liu2024Skill}, which gradually trains the robot to master whole-body behaviors, progressing from simple to challenging terrains.

These terrains include randomly undulations, slopes, discrete obstacles, and stairs. Unlike locomotion tasks, we remove the terrain traversal update mechanism, and instead apply reward threshold updates. Specifically, we monitor the tracking reward within an episode, and when it reaches the threshold, the robot progresses to the next curriculum level in the subsequent episode; otherwise, it regresses to the previous one. This mechanism effectively maintains good tracking performance while ensuring base stability.

\textbf{Domain randomization:}
To reduce the sim-to-real gap, we randomize robot hardware, control, and environment parameters within specified ranges during training~\cite{wjzamp}. As shown in Table \ref{tab:randomization}. This includes PD gains, link mass, payload mass and position, ground friction, and motor strength. Additionally, an observation delay of 0 to 0.02 seconds is introduced to mitigate the system's non-real-time behavior.
\begin{table}[htbp]
\vspace{-0.5em}
        \renewcommand{\arraystretch}{1.0}
        \caption{Randomized parameters and their ranges.}
        \centering
        \label{tab:randomization}
            \begin{tabular}{@{}llll@{}}
                \toprule \\[-4mm]
                \textbf{Parameters} & \textbf{Range [Min, Max]} & \textbf{Unit} \\  \\[-4mm] \midrule
Leg Joint Stiffness & {[}0.8, 1.2{]}$\times$30             &  -          \\
Leg Joint Damping  & {[}0.8, 1.2{]}$\times$0.8              &  -           \\
Arm Joint Stiffness & {[}0.8, 1.2{]}$\times$20             &  -          \\
Arm Joint Damping  & {[}0.8, 1.2{]}$\times$0.5              &  -           \\
Initial Joint Positions  & {[}0.5, 1.5{]}$\times$nominal value  &  rad         \\
Link Mass       & {[}0.8, 1.2{]}$\times$nominal value  &  Kg          \\
Payload Mass    & {[}0, 2{]}                           &  Kg          \\
Payload Position& {[}-0.05, 0.05{]} relative to base origin &  m       \\
Ground Friction & {[}0.05, 2.0{]}                     &  -          \\
Motor Strength  & {[}0.8, 1.2{]}                       &  -          \\
Observation Delay  & {[}0, 0.02{]}                       &  s          \\
[1mm]  \bottomrule
\end{tabular}
\vspace{-12pt}
\end{table}

\subsection{Network}\label{NETWORK}
\textbf{Trajectory-velocity prediction policy network:}
We propose a \textbf{T}rajectory-\textbf{V}elocity \textbf{P}rediction policy network (TVP) consisting of an NAE, a memory encoder, an estimator and a low-level network. The historical trajectories are processed by the NAE to first obtain the NAE embeddings and then predict future poses. The historical proprioceptive state sequence is fed into the memory encoder (ME) to produce the ME embeddings. The ME embeddings are used as inputs to the estimator, which predicts the body’s linear velocity. The predicted future poses, the NAE embeddings, the predicted linear velocity, the ME embeddings, the current proprioceptive state and the last action are fed into a low-level network to generate the joint action.

A trajectory encoder $E_{\theta _{1}}$ extracts features from the historical poses. Then, these features are fed into a decoder $D_{\theta _{2}}$ to predict the subsequent trajectory sequence, as shown in (\ref{trajectory encoder decoder}). This prediction is essential, particularly in teleoperation, where only historical trajectory records are available. Alternatively, when deploying the strategy that utilize DP, the future trajectory can be provided directly, and the TVP will be skipped.
\begin{equation} \label{trajectory encoder decoder}
\begin{aligned}
&\hat{z}_{t}^{tcp}  = E_{\theta _{1}} \left ( o_{t-3:t}^{tcp} \right ) 
&\hat{o}_{t+1:t+4}^{tcp}  = D_{\theta _{2}} \left ( \hat{z}_{t}^{tcp} \right )
\end{aligned}
\end{equation}

We introduce a normalized autoencoder (NAE) to formulate the structure and loss function for the trajectory predictor. The reconstruction loss measures how well the NAE reconstructs the future trajectory from the embeddings $\hat{z}_{t}^{tcp}$:

\begin{equation} \label{NAE loss}
\begin{aligned}
\mathcal{L}_{NAE} &= \mathcal{L}_{rec} = \mathrm {MSE} \left ( \hat{o}_{t+1:t+4}^{tcp} , o_{t+1:t+4}^{tcp}\right )
\end{aligned}
\end{equation}
Specifically, the MSE is used to minimize the discrepancy between the predicted future trajectory and the ground truth. Importantly, we apply the $tanh$ activation function to the encoded embeddings for normalization, constraining them within a limited range. This helps prevent outliers that lie outside the expected distribution, thus aiding in stabilizing the training process. Additionally, we update the trajectory predictor, concurrently with other networks. This both facilitates accurate prediction of future trajectories and provides useful embeddings for the low-level network, thereby promoting the learning of the desired action.

The memory encoder encodes the last five frames of proprioceptive observations. And an estimator is trained through supervised learning to estimate the body’s linear velocity:
\begin{equation} \label{prop velocity estimator}
\begin{aligned}
&\hat{z}_{t}^{prop}  = E_{\theta _{3}} \left ( o_{t-4:t} \right ) 
&\hat{v} _{t}^{p}  = EST_{\theta _{4}} \left ( \hat{z}_{t}^{prop} \right )\\
&\mathcal{L}_{est} = \mathrm {MSE} \left ( \hat{v} _{t}^{p} , {v} _{t}^{p}\right )
\end{aligned}
\end{equation}
The estimation is important because loco-manipulation tasks involve both long-range movement and short-range local operations. The estimated linear velocity of the body effectively captures the distinction between these two types of tasks, enabling more accurate execution of these different loco-manipulation tasks. The encoded embeddings and predicted values, together with current proprioceptive state and the last action, are fed into a low level network, which then outputs the mean vector $\boldsymbol{\mu}_t^b \in {\mathbb{R}^{18}}$ of a Gaussian distribution $\boldsymbol{a}_t^b \sim \mathcal{N}\left(\boldsymbol{\mu}_t^b, \boldsymbol{\sigma}^b \right)$, where $\boldsymbol{\sigma}^b \in {\mathbb{R}^{18}}$ represents the variance of the action, optimized by proximal policy optimization (PPO).

\textbf{Critic network:} 
The critic network comprises a trajectory encoder $E_{\psi_{1}}$ and a height encoder $E_{\psi_{2}}$, which perform dimensionality reduction and representation. $\psi_{1}$ and $\psi_{2}$ represent the critic network parameters. These encoded features, combined with proprioceptive observations, last action and privilege information are used to compute the state value function $V_{t}$, assessing the expected reward for the current state:
\begin{equation} \label{Critic network}
\begin{aligned}
V_{t} = E_{\psi_{3}} \left (E_{\psi_{1}}(o_{t-3:t+4}^{tcp}) \oplus E_{\psi_{2}}(s_{t}^{e}) \oplus o_{t} \oplus a_{t-1} \oplus s_{t}^{p} \right )
\end{aligned}
\end{equation}

\subsection{Trajectory Generation and Sampling}
Trajectory generation encompasses both training and deployment, with the subsequent sections detailing the corresponding generation and sampling strategies.

\textbf{Trajectory library in training:} 
We introduce a trajectory library containing real-world TCP poses from multiple manipulation tasks. During training, the initial points of these trajectories are set as the end-effector poses following episodes reset, preventing low-quality samples in the early stages.

We select trajectories from several typical tasks for learning, such as `clean table', `pour coke', `open container', `pick bear', `pick cup', and `unplug charger'~\cite{Zhaxizhuoma2025FastUMI}. Each task includes 200 trajectories. More trajectories can be added to expand multi-task learning. We also incorporate `pushing' from~\cite{Huy2024UMIonLegs}. These tasks cover repetitive motions, pick-and-place, rotational opening and closing, and long-distance movements, making them representative of common real-world manipulation tasks. The data is stored as separate pickle files. Furthermore, guided by the tracking reward, the robot demonstrates a certain ability to generalize to previously unseen trajectories.

\textbf{Trajectory sampling update:}\label{Trajectory_Sampling_Update}
The complexity of task trajectories varies: some involve more changes in both position and orientation, while others involve fewer changes. If trajectories are sampled uniformly across all tasks, the policy may tend to focus on tracking easier trajectories, as they lead to higher rewards, while performing poorly on more challenging task. This can result in the policy getting trapped in local optima.

To address this, we propose an adaptive trajectory sampling and reward factor updating mechanism, which differs from the uniform sampling throughout the whole training. In the initial stage, trajectories from all tasks are sampled with equal probability. Meanwhile, the average tracking rewards for each task are recorded. When a task’s tracking reward changes, its sampling probability is updated accordingly as follows, prioritizing more challenging trajectories to enhance learning. 
\begin{equation} \label{trajectory sampling probability}
\begin{aligned}
&{P}'_{n}=\frac{1}{1+\exp\left ( k_n\left ( \bar{r} _{n}^{g} - \lambda_n \right )  \right ) }  \\
&P_{n}=\frac{P'_{n}}{ {\textstyle \sum_{i=1}^{N}}P'_{i}}, i,n = 1,...,N
\end{aligned}
\end{equation}
where $\bar{r} _{n}^{g}$ denotes the average tracking reward for each task trajectory. $\lambda_n$ is the reward performance threshold for task $n$. By tuning the values of $\lambda_n$, we can adjust the sampling probability curves of the reference trajectories of different tasks, so that we can achieve similar reward levels for different tasks. $\lambda_n$ is determined by the user according to reward curves of all tasks after policy training. To prevent forgetting, this mechanism retains a minimum probability for each trajectory.

Meanwhile, each task trajectory adopts its own reward coefficient factor. Specifically, after repeated adjustments, a predefined set of average pose error thresholds and task reward factors is established as $\left \{ \left ( T_{k, n}, \sigma_{k, n} \right ) \right \} $. The average pose error for each task is calculated as follows, including Euclidean distance and angular error.
\begin{equation} \label{average pose error for each task}
\begin{aligned}
&e_{n}^{p} = \Vert\boldsymbol{p}_{n, {\rm tcp}}^{\rm target}-\boldsymbol{p}_{n, {\rm tcp}}\Vert_2  \\
&e_{n}^{o} = { \arccos  (0.5 {\textstyle \sum_{i}}  (\boldsymbol{\rm {R}}_{n, {\rm tcp}}^{\rm target}\cdot\boldsymbol{\rm {R}}_{n, {\rm tcp}}^{T})_{ii}-0.5  )} 
\end{aligned}
\end{equation}
where $\sigma_{n}^{p}$ and $\sigma_{n}^{o}$ are then updated according to (\ref{tracking reward factor update}).
\begin{equation} \label{tracking reward factor update}
\begin{aligned}
&\sigma _{n}^{p} = \begin{cases}
 \sigma _{k,n}^{p} & \text{ if } k=\min\left \{ j:\bar{e}_{n}^{p} < T_{j,n}^{p} \text{exists} \right \}   \\
 \sigma _{max,n}^{p} & \text{ otherwise}
\end{cases}  \\
&\sigma _{n}^{o} = \begin{cases}
 \sigma _{k,n}^{o} & \text{ if } k=\min\left \{ j:\bar{e}_{n}^{o} < T_{j,n}^{o} \text{exists} \right \}   \\
 \sigma _{max,n}^{o} & \text{ otherwise}
\end{cases} 
\end{aligned}
\end{equation}
The factor is adaptively updated based on changes in the reward values for each task trajectory. As the training progresses, this factor increases the tracking requirements for each task.

\textbf{In deployment:}
During deployment, the target trajectory can be provided either by teleoperation or by pre-trained DP.

Using the handheld gripper for teleoperation allows humans to directly control the quadruped robot with an arm, and expands the robot's ability to perform whole-body behaviors. As mentioned earlier, the policy can generalize to track previously unseen trajectories within a reasonable range.
The handheld device uses the RealSense T265 to record poses. At startup, we set the arm’s TCP as the trajectory origin, aligning it with the T265’s initial point. As the operator moves, the recorded poses are fed into the policy for tracking. Meanwhile, GoPro-based marker distance estimation directly triggers gripper commands.

The target trajectory output by the DP is applied to an automated loco-manipulation task. In our case, the DP is a pre-trained module using the training approach in~\cite{Chi2023Diffusion, Zhaxizhuoma2025FastUMI}. In the DP training, a ViT-based CLIP visual encoder~\cite{Radford2021Learning} is employed with ViT Base/16 architecture and an image resolution of 224×224. The policy input includes the arm’s proprioception and fisheye images at each time step, with the target trajectory defining the output.

\section{Experiments and Results}
\subsection{System Setup}
We use Isaac Gym to collect data from 4,096 agents for whole-body policy training. The policy is updated every 50 simulation steps and trained on an NVIDIA RTX 3090Ti, requiring about 8 hours for 10,000 iterations.

\begin{figure}[htbp]
\includegraphics[width=0.9\linewidth]{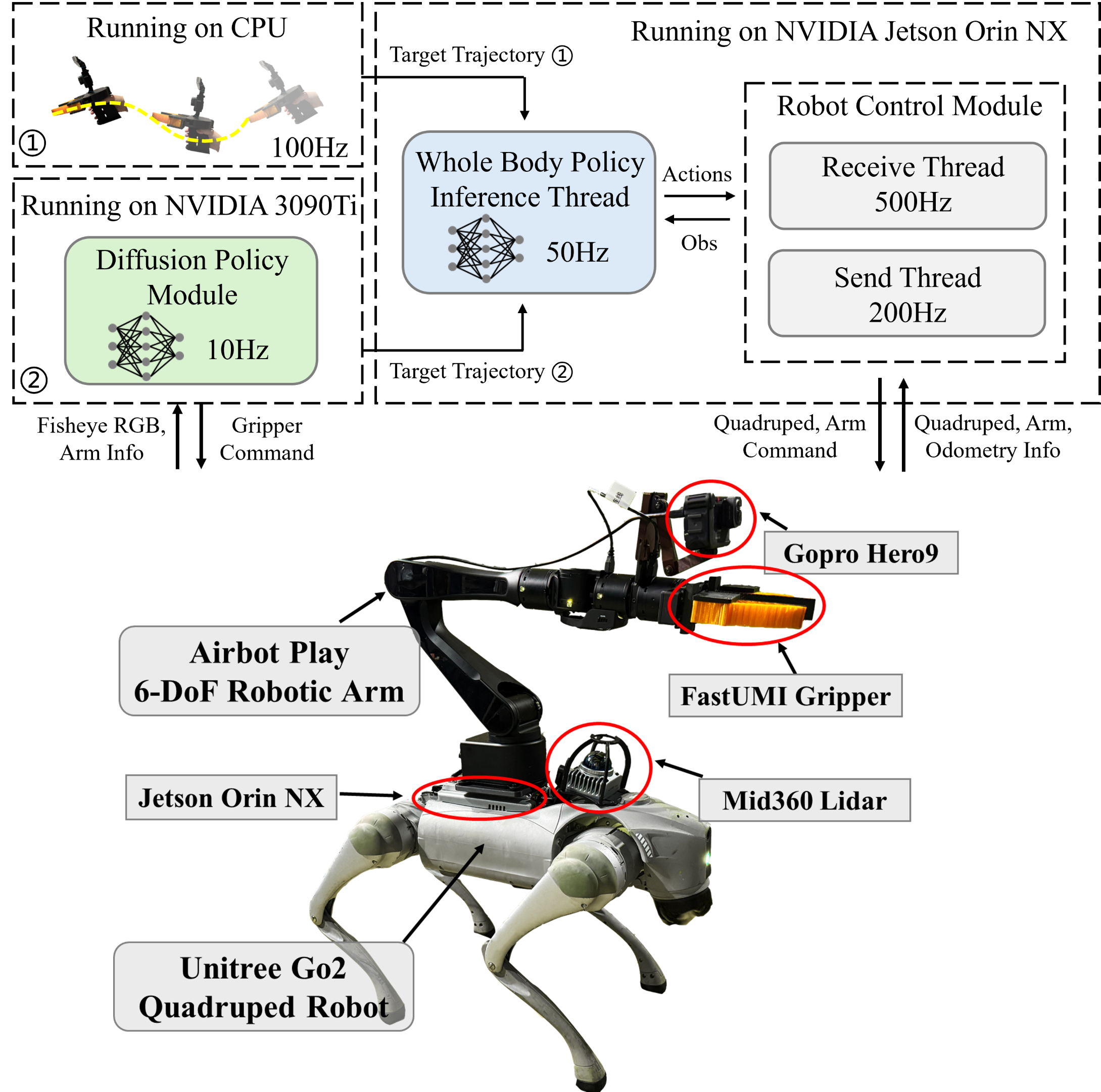}
\centering
\caption{A multi-computer, multi-thread system deployment architecture. The host computer captures the handheld gripper's trajectory (\SI{100}{Hz}) or runs the DP (\SI{10}{Hz}). The trajectory is sent to the lower computer via Ethernet. The Jetson Orin NX, mounted on the robot, handles multi-thread tasks, including receiving robot states (\SI{500}{Hz}), whole-body policy inference (\SI{50}{Hz}), and sending control commands (\SI{200}{Hz}).}
\label{fig:asynchronous_communication}
\vspace{-12pt}
\end{figure}

As shown in Figure \ref{fig:asynchronous_communication}, the controller is deployed on a 12-DoF Uniree Go2 quadruped robot with a 6-DoF Airbot Play robotic arm. The arm's end-effector is fitted with a single DoF parallel gripper~\cite{Zhaxizhuoma2025FastUMI}. The robot total weights \SI{19.9}{\kilogram}. Both the arm and the quadruped robot are powered by a battery on Go2. We implement asynchronous communication for our system through multi-thread. The whole-body policy runs onboard the Go2’s NVIDIA Jetson Orin NX at \SI{50}{Hz}. Inference of DP and the handheld gripper run on a PC with an NVIDIA RTX 3090 Ti that transmits the target trajectory via Ethernet.

During deployment, it is necessary to obtain the target and end-effector poses in the world frame. A Livox Mid-360 LiDAR is mounted on Go2, and an odometry~\cite{Chen2023Direct} is used to build the world frame, with the LiDAR’s initial position as the origin. During inference, the poses are calculated through forward kinematics and transformed into the world frame.

\subsection{Simulation Studies}
We demonstrate the necessity and effectiveness of the proposed method through simulation experiments. For experiments with the same setup, we executed 20 trials with different seeds.

\begin{figure}[!htb]
  \vspace{-10pt}
  \centering
  \subfloat{\includegraphics[width = 0.5\hsize]{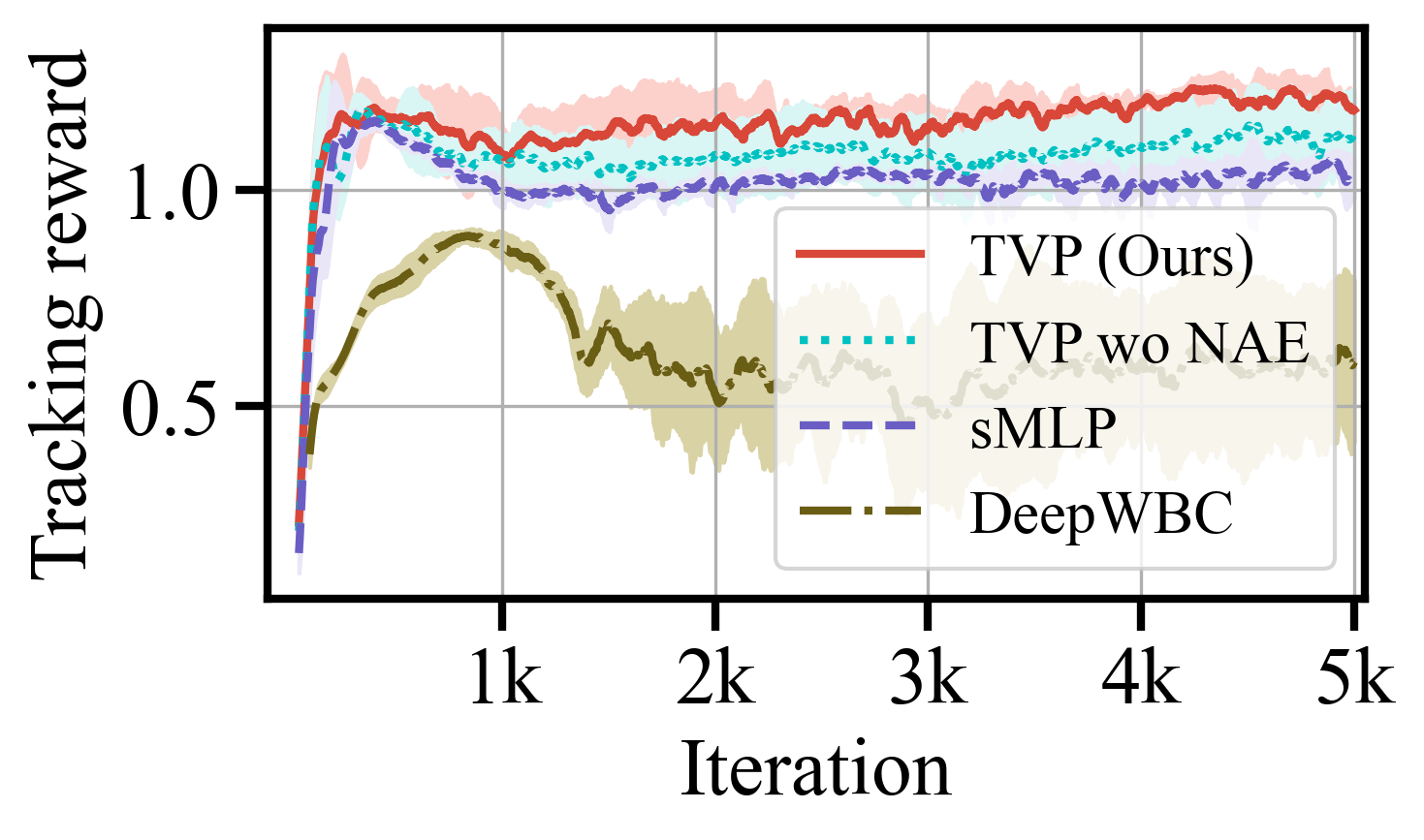}}
  \hspace{-0.15cm}
  \subfloat{\includegraphics[width = 0.5\hsize]{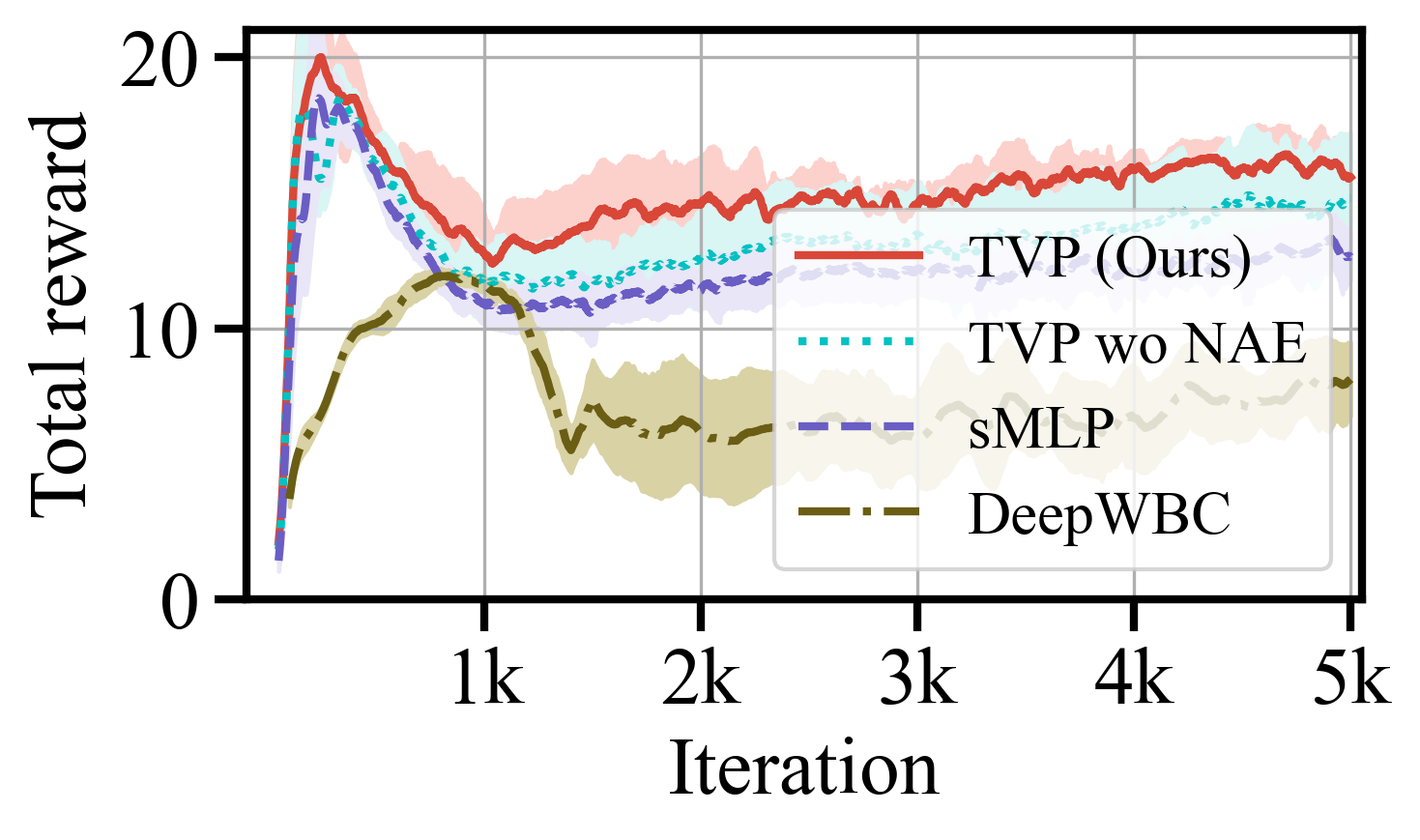}}
  \caption{The trajectory tracking reward (left) and total reward (right) curve during the training of four policy networks: TVP (Ours), TVP wo NAE, single MLP, and DeepWBC.}
  \label{fig:trajectory_tracking_reward_curve}
  \vspace{-10pt}
\end{figure}

\begin{figure}[!htb]
\vspace{-10pt}
\includegraphics[width=0.96\linewidth]{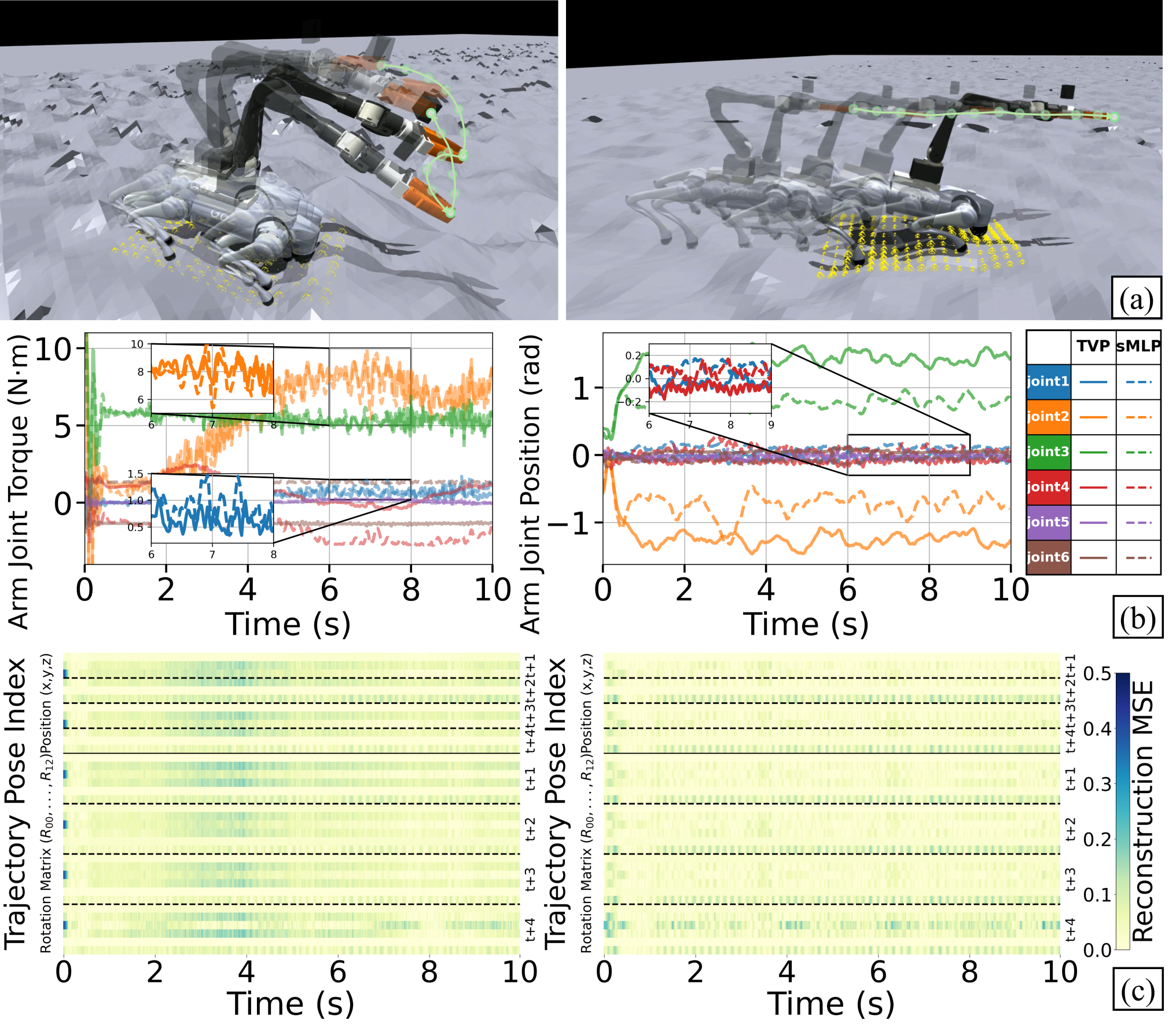}
\centering
\caption{Policy network ablation analysis: \textbf{(a)} The `unplug charger' and `pushing' trajectories in simulation. \textbf{(b)} Arm joint torque and position curves for TVP and sMLP during the tasks, with sMLP showing more jitter. \textbf{(c)} MSE heatmap of future trajectory reconstruction by NAE for both tasks.}
\label{fig:rollout_vs}
\vspace{-10pt}
\end{figure}

\textbf{Policy network wo TVP: }
Using a single MLP (sMLP) actor from the state-of-the-art work~\cite{Huy2024UMIonLegs} as a representative baseline, which has demonstrated good performance on pushing and cup-picking tasks, we highlight the necessity of our TVP policy network for control. This baseline policy does not predict the base velocity. The historical and current trajectories, the current proprioceptive state and the last action are fed into a single MLP network to generate the joint action. Our results show that the baseline policy reward exhibits slower growth and achieves a lower level of convergence, as shown in Figure \ref{fig:trajectory_tracking_reward_curve}. Evaluation further shows that this baseline policy causes larger joint torque and position fluctuations, leading to more noticeable joint jitter, as shown in Figure \ref{fig:rollout_vs}(a) and (b). 
To further quantify the jitter differences, we introduce a mean squared derivative (MSD) of arm joint torques over a time interval, $MSD=\frac{1}{(t_1-t_0)/dt}  {\textstyle \sum_{t=t_0}^{t=t_1-dt}(\tau_{t+dt}-\tau_{t})^{2} }$. In our case, the time interval is from $t_0=5s$ to $t_1=10s$. The results of `unplug
charger' in Table~\ref{tab:msd} show that the TVP can achieve smoother torque control, as it has the lower MSD than sMLP.

\begin{table}[htbp]
\centering
\renewcommand{\arraystretch}{1.0}
\caption{MSD of six joint torques.}
\label{tab:msd}
\begin{tabular}{lcccccc}
\toprule
Method & Joint 1 & Joint 2 & Joint 3 & Joint 4 & Joint 5 & Joint 6 \\
\midrule
TVP   & \textbf{0.0254} & \textbf{0.2660} & \textbf{0.2144} & \textbf{0.0021} & \textbf{0.0026} & \textbf{0.0029} \\
sMLP  & 0.0515 & 0.2867 & 0.2500 & 0.0098 & 0.0038 & 0.0039 \\
\bottomrule
\end{tabular}
\vspace{-12pt}
\end{table}

\textbf{TVP policy network wo NAE: }
In teleoperation, where the future trajectory is unobservable, we conducted an ablation study to evaluate the NAE, which predicts future trajectories from past data. In this ablation setting, the historical proprioceptive state sequence is fed into the ME to produce the ME embeddings. The ME embeddings are used as inputs to the estimator, which predicts the body’s linear velocity. The predicted linear velocity, the ME embeddings, the historical and current trajectories, the current proprioceptive state, and the last action are fed into a low-level network to generate the joint action.
As shown in Figure \ref{fig:trajectory_tracking_reward_curve}, without trajectory prediction, the policy lacks guidance on “where to go next”, leading to degraded tracking performance: the tracking reward remains lower than that of the NAE-based policy.
Figure \ref{fig:rollout_vs}(c) shows the MSE between the NAE-predicted trajectory and the ground truth. The reconstruction errors of position and the first two pose-matrix rows over four frames are visualized as a heatmap. The results demonstrate that the reconstruction error is minimal, indicating that the NAE embeddings contain information essential for predicting the future trajectory.

Table \ref{tab:trajectory_sampling_tracking_accuracy} shows the policy evaluation for seven tasks. The position and orientation errors of the end-effector’s trajectory tracking were recorded over 800 time steps. Position error is measured using Euclidean distance (cm), while orientation error refers to the minimum rotational angle (radian). The average error over all time steps was computed to assess overall tracking performance. Results show that the TVP, with its future trajectory prediction mechanism, outperforms other policies. The sMLP policy performs the worst due to its limited information extraction and representation ability. The error sources could include the dynamics and noise of the command trajectory, the dynamics of the robots, and the policy network.

\textbf{Policy wo NAE, task-frame target and trajectory input: }
We also add a comparison with DeepWBC~\cite{Fu2023DeepWholeBody}. DeepWBC is obtained as follows. We remove the NAE that predicts future trajectories and discard trajectories as the policy targets, relying instead on a single target point. The end-effector pose is defined in the body frame. Through the above setting, we can obtain a method similar to DeepWBC.
The experimental results for each task are presented in Table \ref{tab:trajectory_sampling_tracking_accuracy}. Because the target is defined in the world frame rather than the body frame, our policy is not sensitive to body motion. Because our method has continuous trajectories as reference targets and has a prediction mechanism, our policy’s tracking performance is better than that of DeepWBC.

\begin{figure}[htbp]
\vspace{-0.5em}
  \centering
  \subfloat{\includegraphics[width = 0.8\hsize]{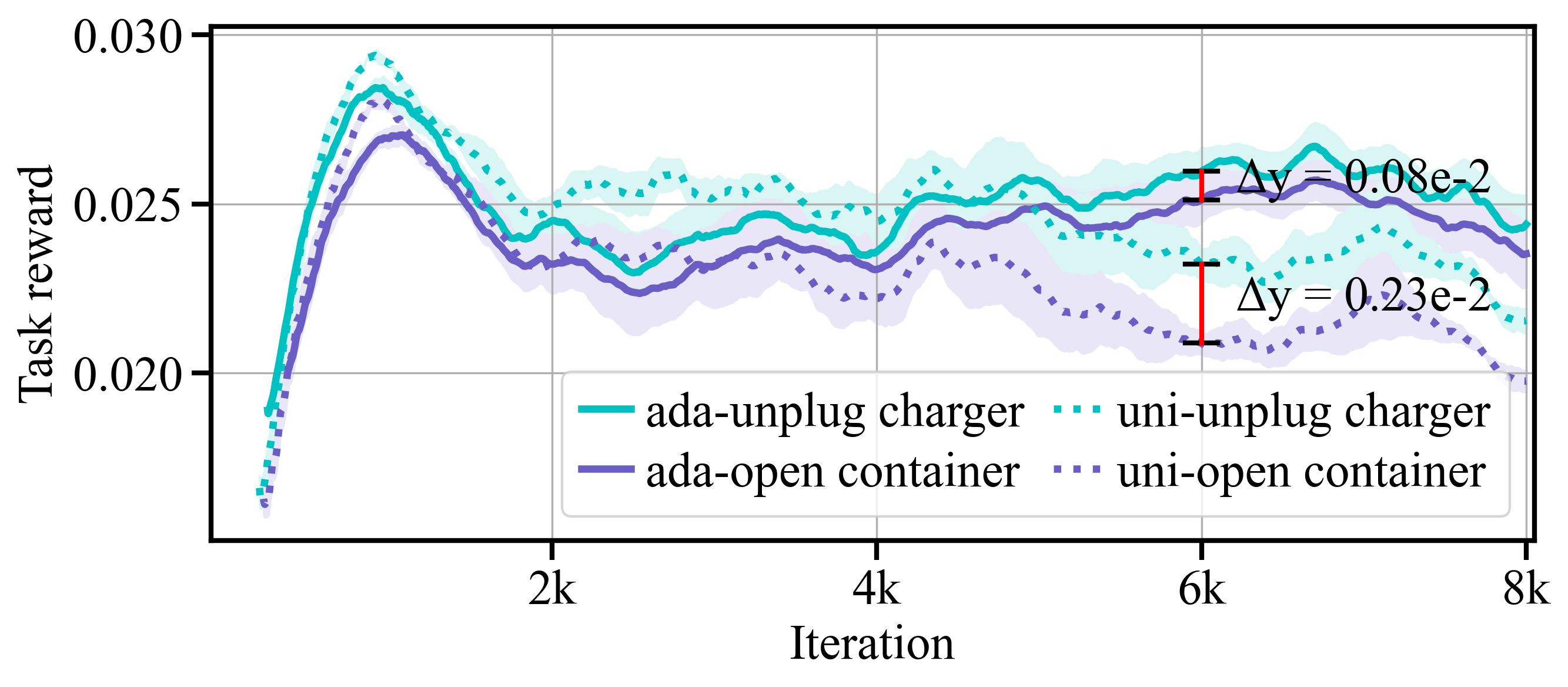}}
  \caption{The trajectory tracking reward for two tasks under adaptive sampling and uniformly sampling.}
  \label{fig:sampling_reward_curve}
\vspace{-0.5em}
\end{figure}

\textbf{Adaptive trajectory sampling: }
To assess the necessity of adaptive trajectory sampling mechanism, we set all robots to uniformly sample (Uni-sample) task trajectories throughout the training. 
We record the reward curves for two typical tasks during training, as shown in Figure \ref{fig:sampling_reward_curve}. The results show that uniform sampling cannot effectively balance task performance. Initially, reward curves for tasks with different pose variations are close. Over time, the rewards diverge. Tasks with more pose variation converge to lower reward due to insufficient attention. In contrast, adaptive sampling maintains similar rewards across tasks, balancing tracking performance. Table \ref{tab:trajectory_sampling_tracking_accuracy} indicates that tasks with frequent trajectory changes (e.g., "open container") see a significant decline in tracking accuracy.

Moreover, the TVP (Ours) and the TVP wo NAE have lower variances than the baselines (sMLP and DeepWBC). Although the TVP (Ours) has a similar variance to the TVP wo NAE, the TVP (Ours) has the lowest mean tracking errors across all tasks.

\textbf{Impact of prediction errors: }
To evaluate the impact of prediction errors on policy performance, we inject Gaussian noise with increasing magnitudes into both the position (1 cm, 2 cm, 10 cm) and orientation (0.05 rad, 0.1 rad, 0.5 rad) components of the predicted trajectory. The experimental results are reported in Table \ref{tab:trajectory_sampling_tracking_accuracy}. As the prediction errors grow, the tracking errors of the loco-manipulation also rise, but the increase remains acceptable, and the tasks can still be completed successfully. In extreme cases, where the predictor outputs unreasonable future steps (e.g., Gaussian noise with a mean of 10 cm and 0.5 rad), the loco-manipulation still functions. The reason is that our method does not rely solely on the predicted trajectory but also on the historical ground-truth trajectory, which provides good robustness against prediction errors.

\begin{table*}[htbp]
\vspace{-1.0em}
    \renewcommand{\arraystretch}{1.0}
    \caption{Simulation ablation study: mean 6D tracking errors for each task, and the variance of those mean errors across all tasks. Metrics: Euclidean distance / rotation-matrix angular difference. (cm / radian)}
    \centering
    \label{tab:trajectory_sampling_tracking_accuracy}
    \begin{tabular}{@{}c|ccccccc|c@{}}
    \toprule
        ~ & pushing & unplug charger & pick bear & pick cup & open container & clean table & pour coke & Variance \\ \midrule
        TVP (Ours) & \textbf{1.32}/\textbf{0.08} & \textbf{1.08}/\textbf{0.07} & \textbf{0.87}/\textbf{0.11} & \textbf{0.66}/\textbf{0.05} & \textbf{0.74}/0.06 & \textbf{0.94}/\textbf{0.09} & \textbf{1.07}/\textbf{0.15} & \textbf{0.050}/\textbf{0.0012} \\ 
        TVP wo NAE & 1.78/0.11 & 1.49/0.07 & 1.30/0.13 & 1.26/0.07 & 1.43/\textbf{0.05} & 1.31/0.10 & 1.78/0.22 & 0.049/0.0032 \\ 
        sMLP \cite{Huy2024UMIonLegs} & 2.23/0.12 & 2.02/0.14 & 2.31/0.37 & 2.24/0.14 & 2.20/0.13 & 2.17/0.22 & 3.13/0.54 & 0.133/0.0256 \\ 
        DeepWBC \cite{Fu2023DeepWholeBody} & 8.32/0.51 & 8.05/0.21 & 6.83/0.49 & 7.35/0.21 & 7.09/0.23 & 9.11/0.30 & 7.22/0.61 & 0.663/0.0279 \\
        \midrule
        Noise ($N(1,0.2^2)/N(0.05,0.01^2)$) & 1.43/0.10 & 1.23/0.08 & 0.92/0.11 & 0.79/0.05 & 0.91/0.06 & 1.08/0.09 & 1.26/0.16 & 0.053/0.0013 \\ 
        Noise ($N(2,0.4^2)/N(0.1,0.02^2)$) & 1.59/0.11 & 1.32/0.08 & 0.99/0.12 & 0.88/0.05 & 1.07/0.06 & 1.22/0.10 & 1.33/0.17 & 0.058/0.0016 \\ 
        Noise ($N(10,2^2)/N(0.5,0.1^2)$) & 2.19/0.13 & 1.51/0.11 & 1.28/0.15 & 1.50/0.07 & 1.40/0.09 & 1.39/0.11 & 1.79/0.25 & 0.098/0.0035 \\ 
        \midrule
        Uni-sample & 1.30/0.07 & 1.11/0.08 & 0.96/0.13 & 0.79/0.06 & 1.69/0.10 & 1.18/0.11 & 1.41/0.21 & 0.088/0.0026 \\
        \bottomrule
    \end{tabular}
\vspace{-15pt}
\end{table*}

\subsection{Real-world Experiments}
We further validate the feasibility and effectiveness of our method through real-world experiments, including teleoperation and DP inference. All tasks are executed with a unified whole-body policy, showing strong performance across tasks and enabling zero-shot transfer from simulation to reality. Please refer to our videos in the supplemental materials.

\textbf{Handheld gripper teleoperation: }
Figure \ref{fig:real_robot} I--V showcase real-world demonstrations of whole-body loco-manipulation via teleoperation, including cleaning papers, picking up trash, storing a toy, unplugging a charger, and removing a bolt to open a door. Figure \ref{fig:real_robot} III demonstrates a long-range task. For ground-level tasks, such as picking up trash, the quadruped robot assists the arm by leaning downward, enabling the end-effector to reach lower positions. Tasks well above body height require lifting the body. When the target position lies beyond the current workspace, the quadruped robot leverages its locomotion to reposition itself. These behaviors demonstrate exceptional whole-body coordination capabilities. Two tasks in Table \ref{tab:real_exp} show that the real-world deployment achieves tracking performance comparable to that of the simulation.

\begin{table}[htbp]
    \renewcommand{\arraystretch}{1.0}
    \caption{Real-world experiment results. (cm / radian / \%)}
    \centering
    \label{tab:real_exp}
    \begin{tabular}{@{}cc|ccc@{}}
    \toprule
        ~ & ~ & EE pos err & EE rot err & Success Rate \\ \midrule
        \multirow{2}{*}{Teleoperation}& unplug charger & 1.39 & 0.054 & 98 \\ 
        ~ & open container & 1.02 & 0.047 & 100 \\ \midrule
        \multirow{2}{*}{DP} & unplug charger & 1.37 & 0.039 & 80 \\
        ~ & open container & 1.24 & 0.036 & 85 \\
        \bottomrule
    \end{tabular}
\vspace{-19pt}
\end{table}

\textbf{Inference through DP: }
As shown in Figure \ref{fig:real_robot} VI--VII, we demonstrate the ability to autonomously execute tasks through a pre-trained DP. Specifically, we input the information from a GoPro camera mounted on the arm's wrist, along with the arm's states, into the DP. The DP then generates the trajectory to the target object. The RL whole-body policy is then used to control the joints for execution. Its performance is mainly limited by the results of DP inference. The experimental results indicate that our method can leverage the DP to plan and autonomously perform multi-task operations, as shown in Table \ref{tab:real_exp}.

\section{Conclusion}
We proposed MLM, a reinforcement learning framework that leveraged both real-world and simulation data to address multi-task whole-body loco-manipulation control for quadruped robots. 
To balance performance across multiple tasks, we integrate a data library with an adaptive sampling mechanism.
For teleoperation where only historical trajectories are observable, we introduce a Trajectory-Velocity Prediction policy network to predict future trajectories, enhancing execution performance. Velocity estimation supports task execution across different spatial ranges.
Simulations and real-world experiments on Go2 and Airbot Play also demonstrate the effectiveness of the proposed method.
Looking ahead, our method can be extended to more tasks in new environments via higher-level decision-making (e.g., vision--language models). Moreover, handling highly dynamic tasks may benefit from faster sensor update rates and better actuation response capabilities.

\bibliographystyle{IEEEtran}
\bibliography{./main}\ 

\vfill
\end{document}